\documentclass[conference]{IEEEtran}
\IEEEoverridecommandlockouts
\usepackage[switch]{lineno}
\usepackage{cite}
\usepackage{amsmath,amssymb,amsfonts}
\usepackage[noend]{algpseudocode}
\usepackage{listings}
\usepackage{graphicx}
\usepackage{textcomp}
\usepackage{xcolor}
\usepackage{lineno,hyperref}
\usepackage{subcaption}
\usepackage{array}
\usepackage{makecell}
\usepackage{multirow}
\usepackage{hyperref}
\usepackage[normalem]{ulem}
\usepackage{algorithm}
\usepackage{algpseudocode}
\usepackage{verbatim} 
\usepackage{tikz}
\usepackage{eso-pic}
\usepackage{amsmath}
\usepackage[english]{babel}
\usepackage{amsthm}

\DeclareMathOperator*{\argminA}{arg\,min} 

\def\BibTeX{{\rm B\kern-.05em{\sc i\kern-.025em b}\kern-.08em
    T\kern-.1667em\lower.7ex\hbox{E}\kern-.125emX}}
\begin{document}

\title{Surrogate-Assisted Evolutionary Generative Design Of Breakwaters Using Deep Convolutional Networks}

\makeatletter
\newcommand{\linebreakand}{%
  \end{@IEEEauthorhalign}
  \hfill\mbox{}\par
  \mbox{}\hfill\begin{@IEEEauthorhalign}
}
\makeatother

\author{\IEEEauthorblockN{Nikita O. Starodubcev}
\IEEEauthorblockA{\textit{Natural Systems Simulation Lab} \\
\textit{ITMO University}\\
Saint-Petersburg, Russia \\
nstarodubtcev@itmo.ru}
\and
\IEEEauthorblockN{Nikolay O. Nikitin}
\IEEEauthorblockA{\textit{Natural Systems Simulation Lab} \\
\textit{ITMO University}\\
Saint-Petersburg, Russia \\
nnikitin@itmo.ru}
\and
\IEEEauthorblockN{Anna V. Kalyuzhnaya}
\IEEEauthorblockA{\textit{Natural Systems Simulation Lab} \\
\textit{ITMO University}\\
Saint-Petersburg, Russia \\}
}

\maketitle
\thispagestyle{plain}
\pagestyle{plain}

\begin{abstract}
In the paper, a multi-objective evolutionary surrogate-assisted approach for the fast and effective generative design of coastal breakwaters is proposed. To approximate the computationally expensive objective functions, the deep convolutional neural network is used as a surrogate model. This model allows optimizing a configuration of breakwaters with a different number of structures and segments. In addition to the surrogate, an assistant model was developed to estimate the confidence of predictions. The proposed approach was tested on the synthetic water area, the SWAN model was used to calculate the wave heights. The experimental results confirm that the proposed approach allows obtaining more effective (less expensive with better protective properties) solutions than non-surrogate approaches for the same time.
\end{abstract}


\section{Introduction}
\label{sec_intro}

The problem of optimal maritime structures design is widely discussed in the literature \cite{goda2010random}. Usually, a design engineer makes a decision based on previous experience and takes into account a set of criteria: cost, the effectiveness of structures, time outlay, etc. These problems are resource- and time-consuming since experts should explore a large number of configurations. To increase efficiency, reduce the cost of construction and operation, as well as to reduce the time spent by expert designers, researchers formulate and solve a multi-objective optimization problem \cite{premkumar2022multi}. Evolutionary algorithms (EA) are widely used for solving such tasks \cite{nikitin2020multi}, \cite{diab2017survey}. 

Since the evaluation of the fitness function can be computationally expensive, the convergence of optimization can require too much time and resources. To decrease the requirements, the lightweight surrogate models that approximate original objective functions can be used \cite{asher2015review}. The combination of surrogate models with EA is often called surrogate-assisted evolutionary algorithms (SaEA).


In this paper, we propose a multi-objective surrogate-assisted evolutionary algorithm for automated breakwaters design. It uses a deep convolutional neural network as a surrogate model for breakwater design. We demonstrate that the proposed approach yields less expensive and more effective solutions than non-surrogate approaches, also it allows achieving a good accuracy of approximation while using a relatively lightweight model. One of the main advantages of this approach is working without any pre-trained part. The model has the ability to train from the data obtained during the optimization process. Also, it takes into account configurations of breakwaters with a variable number of breakwaters and their segments.


\section{Related work}
\label{sec_related}
In the task of the generative design of breakwaters, it is necessary to take into account many factors at once: cost of breakwaters, the height of the waves at the points of interest, etc. Thus, it can be formulated as a multi-objective problem. 
To calculate the height of waves at the target points, there is a need to apply a numerical model for the water area simulation. Usually it is based at partial differential equations. The optimization process requires repeatedly calling this numerical tool.

To reduce the computational cost of optimization, researchers often replace an accurate but computationally expensive physical model with a quickly computable model for approximation of objective function - the so-called surrogate model \cite{palar2019use}. Surrogate modeling is actively used to solve problems from different fields: simulating oil reservoirs for maximizing the total production of oil value and forecasting the most profitable oilfields\cite{navratil2019accelerating}, optimizing of the heat-generating components in small electronic devices for control of temperature field \cite{chen2020heat}, obtaining the hydrodynamic performance indexes of ship hull form for increasing its strength \cite{liu2022multi}. Researchers consider a wide variety of models as surrogates: from classical methods (polynomial regression, kriging, support vector regression) to complex ensemble models, deep neural networks, long-short term memory networks, etc \cite{chugh2019survey}. 

Surrogate models are widely used in coastal engineering to avoid the unnecessary runs of metocean models. In \cite{chen2021using} authors apply a random forest model to predict wave conditions in significant points (buoy locations) across the considered domain (Cornwall, South West UK). The output of the physics-based model (significant wave height, mean wave direction, mean zero-crossing period, and peak wave period) is used as input features to the random forest. The surrogate model was trained on historical data from 1989 to 2009 and evaluated for the following year (2010). This approach showed a good performance, the results were compared with numerical prediction. A similar case was considered in \cite{james2018machine}, researchers utilized multi-layer perceptron based on numerical model outputs to solve the regression task and predict the wave height. Also, a support vector machine was used to predict the wave period. This approach has been tested on Monterey Bay and demonstrated good agreement with a numerical model with a 4,000 times improvement in computational speed. In wave protection structures optimization tasks surrogate modeling is also used. In \cite{xu2021ecological} authors developed computational framework SEO for eco-seawalls design. A multi-objective optimization problem was formulated, in which compromise between engineering cost and ecological protection was considered. To analyze the ecological situation and to study hydrodynamic characteristics of eco-seawalls researchers used SDM (species distribution modeling) and CFD (computational fluid dynamic) models respectively. Authors implement kriging interpolation as a surrogate model to replace the computationally expensive CFD model.

In described approaches the researchers imposed some restrictions on the problem. As an example, parameters such as wind direction, the number of wave protection structures, and the size of the considered domain were fixed. Some of these limitations are critical for the generative design of breakwaters. 
Indeed, to find the most effective configuration of breakwaters, it is necessary to vary their number. This limitation cannot be solved using basic surrogate models like kriging, radial basis functions, etc. For example, a variable number of breakwaters will result in a variable number of inputs for the surrogate model. Thus, in the optimization process surrogate will have to be retrained. The proposed algorithm is aimed aims to solve this problem by representing the breakwaters in the form of a three-dimensional tensor and using a deep convolutional neural network.


\section{Problem statement}
\label{sec_problem}
The general statement of the problem of breakwaters design is to generate structures configuration that provided maximum protection of interesting targets (protection area) and possessed minimal cost. The location of the breakwaters is limited by the water (search) area. In addition, prohibited areas can be located inside the search domain. These are areas in which breakwaters should not be located due to intersection with existing objects in the water area. Figure~\ref{example_bw} shows a configuration of breakwaters in search space. In the search space, there are prohibited areas and structures for protection. 
\begin{figure}[h]
\centerline{\includegraphics[width=9cm]{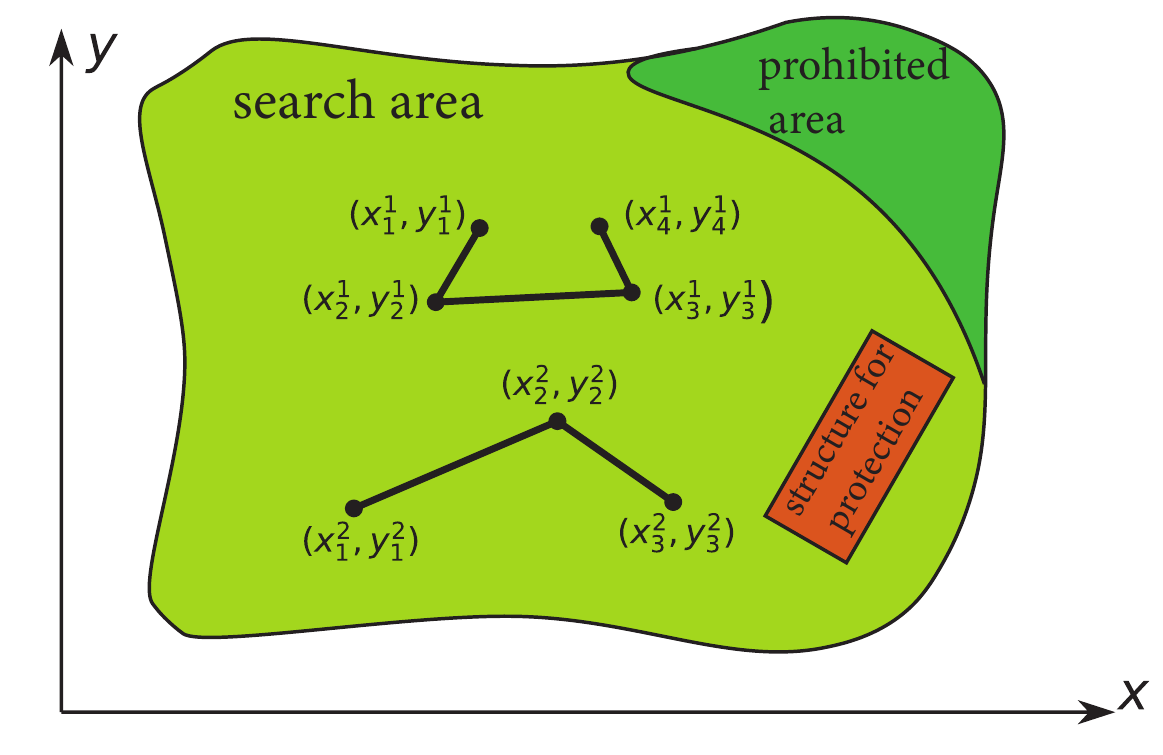}}
\caption{Breakwaters configuration in Cartesian coordinates. Search area (dark green), prohibited area (green) and structure for protection (orange) are defined.}
\label{example_bw}
\end{figure}
The presented configuration include two breakwaters with four and three segments respectively. Structures are parameterized in relative spatial (Cartesian) coordinates, each breakwater node is assigned a spatial coordinate. System of breakwaters can be represented as 
\begin{equation}  
\begin{gathered}
    {\Psi}({\{\xi_{i}^{j}\}}) = {\langle\mathbf{X, Y}\rangle}, \\
\end{gathered}
\end{equation}
with spatial coordinates of nodes for each breakwater in a system (example on Figure ~\ref{example_bw})
\begin{equation}  
\begin{gathered}
	\Psi({\xi}^k_{1}, {\dots}, {\xi}^k_{m}) = (x^k_{1}, y^k_{1}, \dots, x^k_{m}, y^k_{m}) = {\langle\mathbf{x,y}\rangle}^k,
	\label{eq1}
\end{gathered}
\end{equation}	
 where ${\Xi} = {\{\xi_{i}^{j}}\}-$ arbitrary coordinates, and ${\Psi} -$ parameterization function taking values in parametric space ${\Omega}$. There are various parameterization that can be used to represent the structures, Cartesian coordinates just one of them. In some cases, right choice of parameterization can provide improvement in optimization process, in \cite{nikitin2020multi} angular encoding approach demonstrates superiority over relative spatial coordinates. 
 
 So, without loss of generality we may write the optimization problem relating to Cartesian notation in the form
\begin{equation} \label{eq2}
\begin{gathered}
  \langle\mathbf{X^*, Y^*}\rangle=\argminA_{\theta\in\Omega}\mathcal{F}(\langle\mathbf{X, Y}\rangle) \\
  \mathcal{I}_{\Omega^{'}}(\langle\mathbf{X, Y}\rangle) = 0,\\
  \mathcal{D}(\langle\mathbf{X, Y}\rangle,\mathbf{T}) \geqslant \epsilon,
\end{gathered}
\end{equation}
where \( \mathcal{F}(\theta) = (f_{cost}, {f}_{wh}), \mathcal{F}:\Omega \rightarrow \mathbb{R}^n\) \(-\) multi-objective function, \( \mathcal{I}_{\Omega^{'}}\) \(-\) identifier function for location of breakwaters in the forbidden domain \(\Omega^{'} \). Domain \(\Omega^{'} \) consist of prohibited, protection area and outside of search area. \( \mathcal{D}\) \(-\) function for distance from breakwaters to target points \( \mathbf{T}\) and other structures. In our case, number of criteria equals two. Was considered \(f_{cost}({\Psi}({\xi}_{1}, {\dots}, {\xi}_{n})), f_{wh}({\Psi}({\xi}_{1}, {\dots}, {\xi}_{n}))\) \(-\) objective functions correspond to the cost of the breakwater configuration and the wave height at interesting points, respectively. The specific form of these functions can be written by setting the appropriate parameterization. Following~(\ref{eq1}) (Cartesian parameterizaiton):
\begin{equation}
\begin{aligned}
    f_{cost}(\langle\mathbf{X, Y}\rangle) = \sum_{j}\sqrt{\sum_{i} (x_{i}^j - x_{i+1}^j))^2 + (y_{i}^j - y_{i+1}^j))^2}
\end{aligned}
\end{equation}
\(f_{cost}\) defines like sum of Euclidean norms. In our formulation cost of the configuration of breakwaters depends only on their length, indexes \(j, i\) are corresponding to number of breakwater and number of segment, respectively. Surrogate formulation can be denoted for function of wave protection \(f_{wh}\) as follows
\begin{equation}
\begin{aligned}
    f_{wh}(\mathbf{T}) =
                \begin{cases}
    \mathbf{M_{phys}}(\langle\mathbf{X, Y}\rangle,\mathbf{T},\mathbf{H}),   \mathcal{L_{CNN}}(\langle\mathbf{X, Y}\rangle) = 0 \\
    \mathbf{M_{CNN}}(\langle\mathbf{X, Y}\rangle,\mathbf{T},\mathbf{M^{hist}_{phys}}), \mathcal{L_{CNN}}(\langle\mathbf{X, Y}\rangle) = 1\\
                 \end{cases}
\end{aligned}
\end{equation}
 It is difficult to represent the certain mathematical expression for the function \(f_{wh}\), since it is necessary to apply a numerical wave model \(\mathbf{M_{phys}}\) with  \(\mathbf{H}\) simulation parameters (wind direction and speed, water depth, etc) or surrogate model in a form of a deep CNN model  \(\mathbf{M_{CNN}}\) that is trained on accumulated results of wave model \(\mathbf{M^{hist}_{phys}}\). Choice between physical and surrogate models is made by surrogate assistance classification model \(\mathcal{L_{CNN}}\) that predict confidence of \(\mathbf{M_{CNN}}\) model for certain variant of breakwaters system \(\langle\mathbf{X, Y}\rangle\).

\subsubsection*{Hypothesis}\label{hypmo} The surrogate-assisted optimization makes it possible to obtain more efficient (in terms of cost and protection quality) structures compared to non-surrogate approaches for the same time of work.

\section{Multi-objective surrogate-assisted evolutionary approach for breakwaters design}

In the breakwaters optimization task, we have to use numerical models to reproduce the environmental conditions. This leads to the following difficulties: 1) applying gradient-based methods is problematic; 2) numerical modeling is time-consuming. For these reasons, we designed a multi-objective surrogate-assisted evolutionary approach for breakwaters optimization. The different parts of the algorithm will be described below.

\subsection{Evolutionary approach}

Evolutionary algorithms may be used to solving multi-objective optimization problems. The main approach of such algorithms is to approximate the Pareto frontier \cite{zitzler2004tutorial}. 
{\setlength{\parindent}{0cm} Knowledge about Pareto optimal set allows decision-makers to choose the best compromise solution. For solving problem~(\ref{eq2}) methods based on Pareto dominance \cite{zitzler2004tutorial} and diversity preservation \cite{zitzler2001spea2} were implemented.} One of the main part of every evolutionary algorithms is genetic operators. We implemented five different mutations and one crossover operator. It is shown in Figure~\ref{example_mutat}.

\begin{figure}[h]
\centerline{\includegraphics[width=6cm]{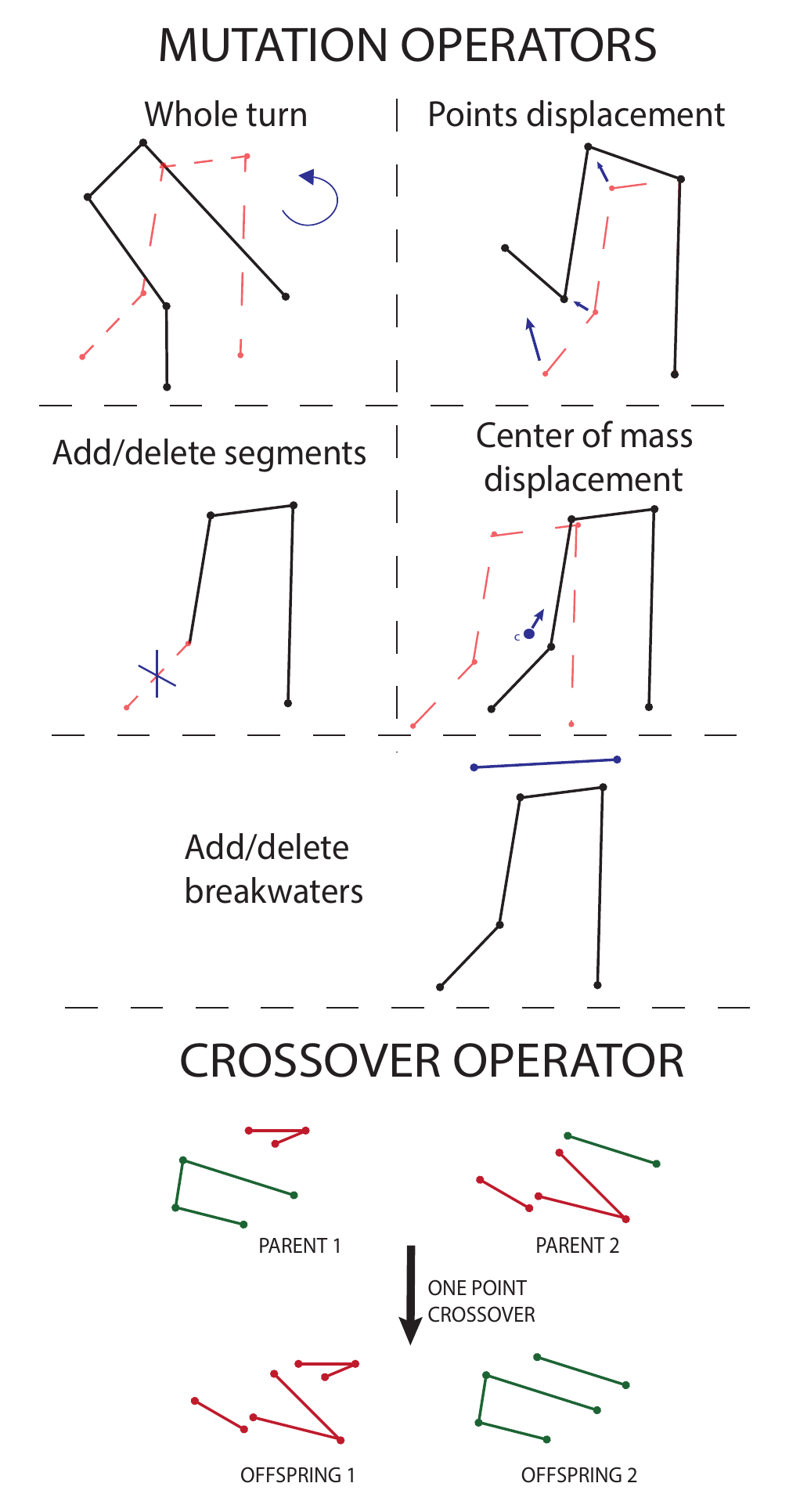}}
\caption{Implemented mutation and crossover operators for breakwaters configuration.}
\label{example_mutat}
\end{figure}
It can be seen that implemented genetic operators have a local nature, which leads to the predominance of exploitation over exploration in a utilized algorithm.
To deal with the constrained in optimization problem~(\ref{eq2}), the genetic operators were applied to individuals until the corresponding transformation satisfies the constraints. This procedure is described in more detail in Alg.~\ref{alg_gp}. As a genetic strategy, \( (\mu,\mu)\) was used, where \(\mu\) - population size before and after crossover and mutation. After the next optimization step, the offspring population entirely substitutes the parent population.

\subsection{Deep CNN as a surrogate model}
 
In the proposed approach, we used a surrogate model that approximate the real numerical model output. One of the main difficulties of this task is that the number of breakwaters and their segments can be different, which causes the different number of variables in Cartesian parameterization presented in Eq.~(\ref{eq1}). To accomplish this, we utilized universal parameterization in the form of a three-dimensional array, called input mask, which does not depend on the number of segments and breakwaters. 
The input mask is an image of the entire water area, instead of wave heights, random values from the normal distribution are used. To work with this parameterization, we used a convolutional neural network. We implemented architecture of deep convolutional neural network based on encoder-decoder structure  \cite{he2016deep}, shown in Figure~\ref{arch_CNN}, for solving regression task. 

The main feature of the surrogate model is that it has two types of outputs. The first is designed to reconstruct the entire field of sea waves in the whole water area using three-dimensional array parameterization, the second obtain the heights of sea waves in important points. The second output is obtained from the first by taking a linear transformation from the intensity of the pixels corresponding to the important points. Only the values from the second output are needed to solve the initial problem and it is possible to think that the first output is useless. The first output is a key factor in the model's understanding of  dependence between wave height in target points and breakwater configuration since input masks have no information about the field of sea waves. 
\begin{figure*}[!h]
\centerline{\includegraphics[width=14cm, height=4cm]{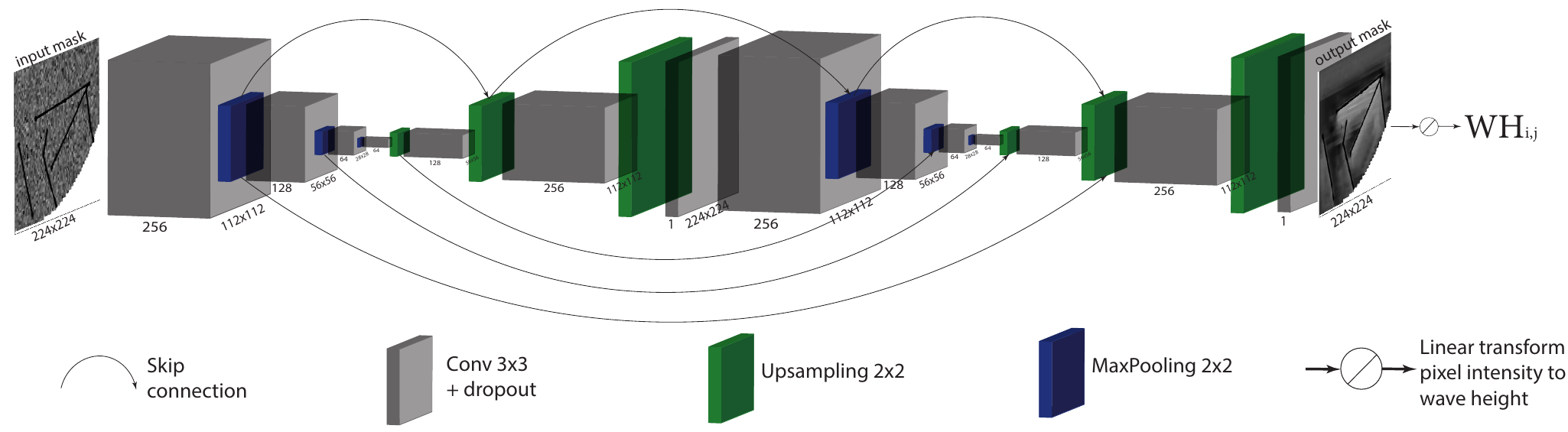}}
\caption{Surrogate model architecture. Outputs of the CNN is an output mask and wave heights at important points. The  model has approximately two million parameters and consists of convolutional layers with 3x3 kernels, upsampling, max pooling, and skip connections}
\label{arch_CNN}
\end{figure*}
\begin{figure*}[!h]
\centerline{\includegraphics[width=18cm, height=17cm]{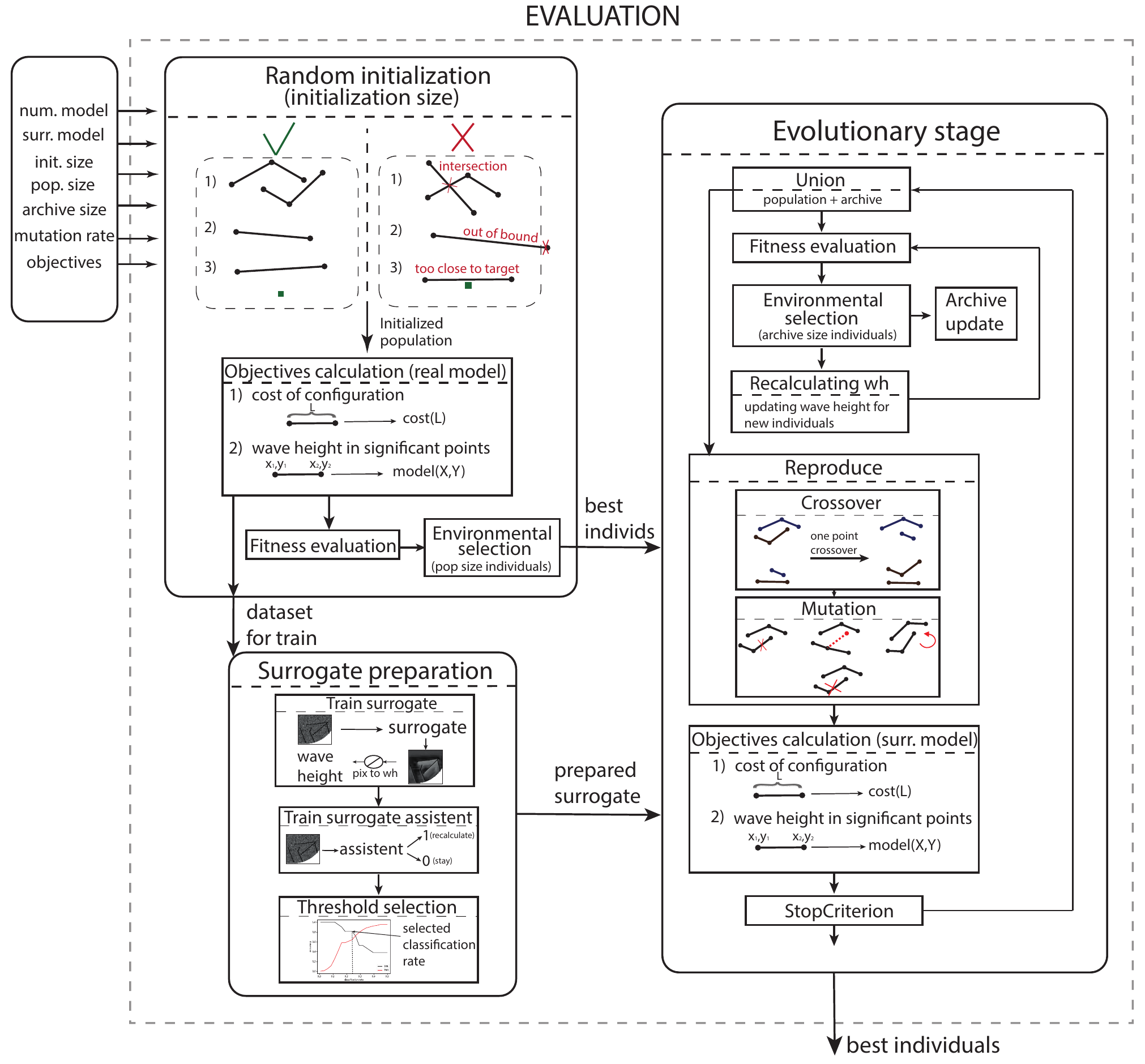}}
\caption{Proposed surrogate-assisted evolutionary approach consisting of three main stages: random initialization, surrogate preparation, evolutionary stage.}
\label{surr}
\end{figure*}

\subsection{Surrogate assistant}

 High accuracy of the surrogate approximation is especially important for the multi-objective optimization task. Even small errors in the estimation of the objective function can cause a redistribution of individuals on the Pareto frontier, which can lead to divergence. Consider the following ways to deal with this issue: 1) develop surrogate models with accuracy close to 100\% (which is almost impossible); 2) develop models that would estimate how confident the surrogate is in its prediction, in cases where the surrogate is uncertain uses a real numerical model; 3) modify the optimization process in such a way that solutions that were erroneously determined by the surrogate as good, could not get into the archive population and do not ruin the convergence process. We developed methods based on the third and second points. A neural network classifier called surrogate assistant was used as a confidence model. The surrogate assistant has a similar structure to the surrogate model and takes three-dimensional parameterization as input.
It solved the binary classification problem, i.e., it determined whether it was necessary to apply a real model to a given configuration of breakwaters. For the classifier, a dataset was labeled, the item was assigned a label 1 (recalculate wave height by real model) if the surrogate error on it was higher than a predefined threshold, otherwise, it was assigned a label 0. The classification process should be considered in more detail. The main task is to define a threshold of the classifier according to which there will be a division into two classes. False-negative errors are more important here since we do not want to preserve individuals with a large surrogate model error in a the population erroneously. But true negative errors are important to ensure that all individuals are not recalculated by the real model. Thus, the threshold was designated based on the analysis of the behavior of true positive (TPR) and true negative rates (TNR) with the variation of the classifier threshold. In this analysis, we focus most on TPR.

Besides, the optimization process was modified in such a way as to individuals that should be included in the archive population after the next optimization step are recalculated by the real model. Such an approach is more time-consuming, but on the other hand, will not allow adding unsuccessful individuals to the archive.

\subsection{Proposed approach}

The proposed multi-objective surrogate-assisted evolutionary approach is shown in Figure ~\ref{surr} and Alg.~\ref{alg_gp}. This approach consists of three main stages called random initialization, surrogate preparation, and evolutionary stage. In the first stage, the population that satisfies the constraints is initialized with initialization size (initialization size much bigger than population size). Objectives functions are calculated for each individual in a large population, after which the prepared dataset is sent to the surrogate preparation stage. Also, individuals are assigned a fitness value and the best individuals (in the amount of pop size) are selected from a large population and sent to an evolutionary stage. The third stage is based on the modified optimization process described above. For the optimization process to be as efficient as possible, the third stage starts its work using a real model. After the surrogate model is ready, this stage continues to work with the surrogate. The proposed stage of initialization of a large population has two advantages: 1) compensation for low exploitation rate at the evolutionary stage; 2) preparing a diverse dataset for training surrogates. The evolutionary state is algorithm-agnostic, so various implementations of multi-objective evolutionary algorithms (e.g., SPEA2 or MOEA-DD) can be used in it.

\begin{algorithm*}
\caption{Proposed surrogate-assisted evolutionary procedure}

\begin{algorithmic}[1]
\Procedure{Optimize}{} 
    \State \underline{Input:} $ CostObjective, SurrModel, RealModel, popSize, archSize, datasetSize, mutatRate, stopCriteria $ 
    \State \underline{Output:} $archPop$
    \State $pop\gets \Call{InitPop}{datasetSize} $
    \State $popWh\gets \Call{RealModel}{pop} $ \Comment{Calculating wave height by real model for large population}
    \State $archPop, archWh\gets \emptyset $
    \State $fit\gets \Call{Fitness}{pop, popWh} $
    \State $pop, popWh\gets \Call{EnvSelection}{pop, fit, popSize} $ \Comment{Selecting best popSize individuals from large population (exploration phase)}
    \State $\Call{SurrModel.Preparation{}} $ \Comment{Preparaining surrogate models}
    \While{$stopCriteria$} \Comment{Starting exploitation phase}
        \State $unionPop, unionWh \gets pop\cup archPop, popWh\cup archWh  $
        \State $unionFit\gets \Call{Fitness}{unionPop, unionWh} $
        \If{$\Call{SurrModel.state}{}{ =ready}$}
            \State $archNew, archNewWh\gets \Call{EnvSelection}{unionPop, unionFit, archSize} $ \Comment{Selection of the fittest individuals}
            \State $archNewWh\gets \Call{RealModel}{archNew} $ \Comment{Recalculating wave height for selected individuals by real numerical model}
            \State $unionFitNew\gets \Call{Fitness}{archPop\cup archNew, archWh\cup archNewWh} $
            \State $archPop, archWh\gets \Call{EnvSelection}{archPop\cup archNew, unionFitNew, archSize} $ \Comment{Additional selection necessary to preserve the archive's elitism}
            \State $\Call{Model}{}\gets \Call{SurrModel}{}$
        \Else
            \State $archPop, archWh\gets \Call{EnvSelection}{unionPop, unionFit, archSize} $
            \State $\Call{Model}{}\gets \Call{RealModel}{}$
        \EndIf
        \State $pop\gets \Call{Reproduce}{unionPop, mutatRate} $
        \State $popWh\gets \Call{Model}{pop} $
        
    \EndWhile\label{constraintwhile}
    \State \textbf{return} $archPop$ 
\EndProcedure \label{optimizer}
\Procedure{Reproduce}{pop, mutatRate} 
    \State $newPop\gets \emptyset $
    \For{$individ1$ \textbf{in} $pop$}
        \For{$individ2$ \textbf{in} $pop$}
            \State $out\gets \Call{Crossover}{individ1, individ2} $
            \If{$\Call{Constraints}{out}$} \Comment{Constraints satisfaction check}
                \State $newPop\gets newPop \cup out$
            \Else
                \State \textbf{continue}
            \EndIf
    \EndFor
        \EndFor
    \State $newPop\gets \Call{Mutation}{newPop, mutatRate} $
    \State \textbf{return} $newPop$ 
\EndProcedure \label{Reproduce}
\end{algorithmic}
\label{alg_gp}
\end{algorithm*}

\section{Experimental studies}

\subsection{Experimental setup}

To conduct the set of experiments, we developed the synthetic setup of the water area. It is shown in Figure~\ref{syntethic_cases}. Bathymetry (water depth in every point of the spatial grid), wind direction, and speed are set to simulate the water area. In this case, the aquatic area includes two targets (the points chosen for protection from sea waves), a land part, and two static non-optimizable breakwaters attached to the shore. The SWAN model was used as a numerical model. 
\begin{figure}[H]
    \centering
    {{\includegraphics[width=8.5cm]{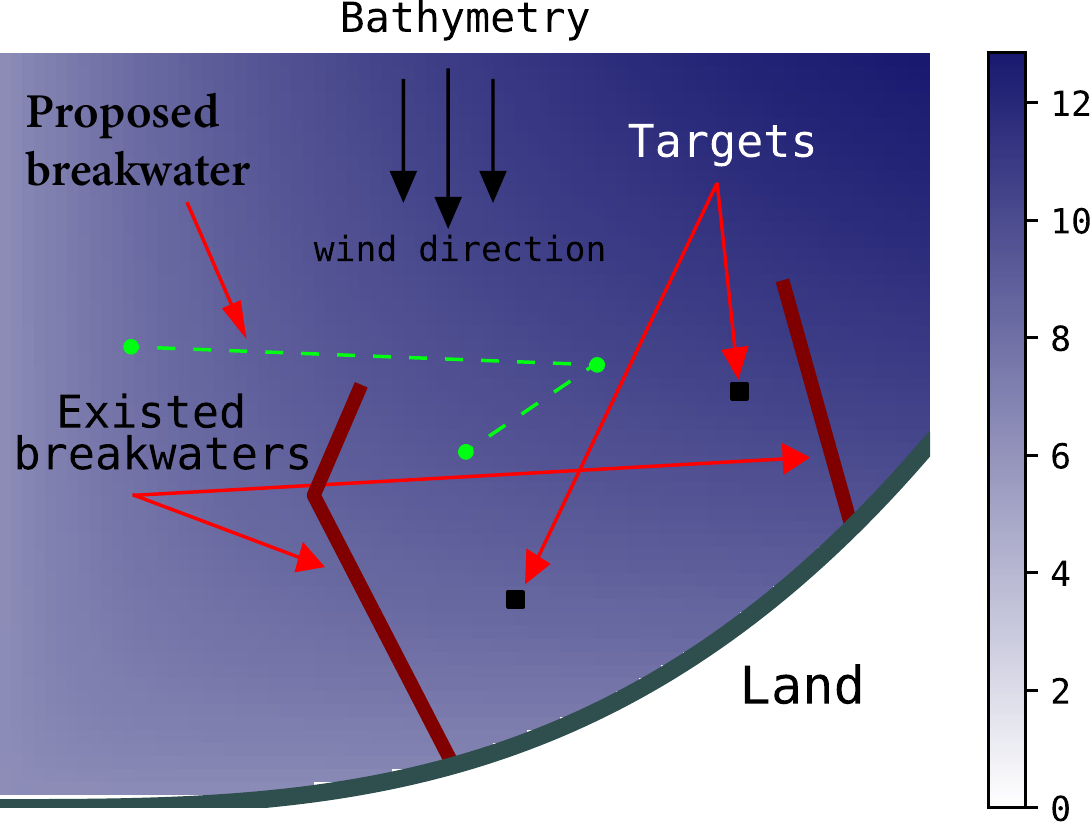} }}%
    \caption{Coastal area with bathymetry increasing from the lower left corner to the upper right. Two fixed breakwaters, land and two important points (targets) are defined.}%
    \label{syntethic_cases}
\end{figure}
\setlength{\parindent}{0cm}Four optimization approaches were compared:
\begin{itemize}
    \item Proposed surrogate-assisted evolutionary approach shown in Figure~\ref{surr};
    \item Same evolutionary approach as proposed, but without a surrogate model (the real model utilizes instead);
    \item Baseline approach. To show the importance of a large exploration phase we considered an approach that uses a small population size at the random initialization stage and uses a real model only during the evolutionary stage;
    \item Random search.
\end{itemize}
SPEA2 algorithm was used as an underlying evolutionary algorithm for both baseline and proposed approach. For the first and second approaches, were fixed following parameters: \(S_{init}\) = 1500 (initialization size), \(S_{pop}\) = 40 (population size), \(S_{arch}\) = 20 (archive size), \(\mu\) = 0.35 (mutation rate), the criterion for stopping the algorithm was achievement for 2500 evaluation of the real model. The total optimization time was limited to approximately ten hours. The convolutional neural network was trained by the ADAM optimizer with batch size equal to 12 and learning rate decreasing from 0.001 with factor 0.95 every 5 epochs. Dataset was split in training and test in a ratio of 80/20. As a loss function cross-entropy and weighted mean absolute error was chosen for the first and second output of the neural network respectively.

\subsection{Results of surrogate modeling}
\setlength{\parindent}{0.3cm}Correspondence between predicted and real wave height in target points is shown in Figure~\ref{wh}. 
\begin{figure}[H]
    \centering
    {{\includegraphics[width=8.5cm]{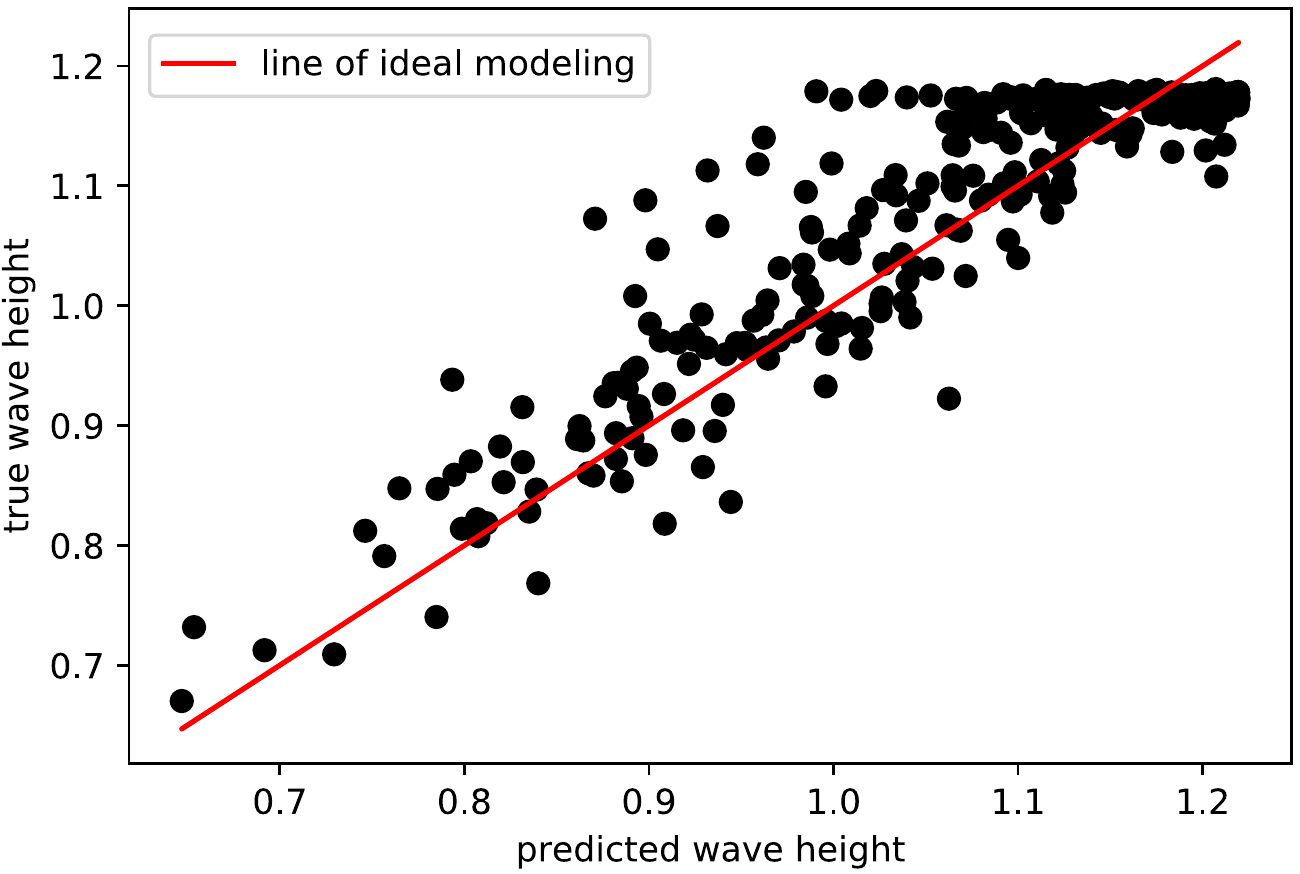} }}%
    \caption{Correspondence between predicted and real wave height.}%
    \label{wh}
\end{figure}
\setlength{\parindent}{0cm}In this case, the following quality of determining the wave height is achieved: mean absolute percentage error (MAPE) - 4.24\%, mean absolute error (MAE) - 0.043. As can be seen from the Figure~\ref{wh}, a lot of predicted values are overestimated. These errors were less critical for multi-objective optimization than underestimating errors, which can lead to divergence of the optimization process. This aspect is related to the Pareto dominance used in the definition of the fitness function, individuals with underestimated objectives dominate other individuals and consequently have better fitness. These mistakes would lead to a violation of elitism in the archive population. The next step following the readiness of the surrogate model is assistant preparation. In this experiment, the dataset was labeled with a threshold error equal to \(5\%\) by MAE. The quality of a classifier was evaluated by the ROC-AUC metric (due to the high imbalance of classes) and is equal to 0.77. The last step of surrogate preparation is the selection of a classification threshold. The analysis of TPR/TNR is presented in Figure~\ref{rates}. \begin{figure}[H]
    \centering
    {{\includegraphics[width=8.5cm]{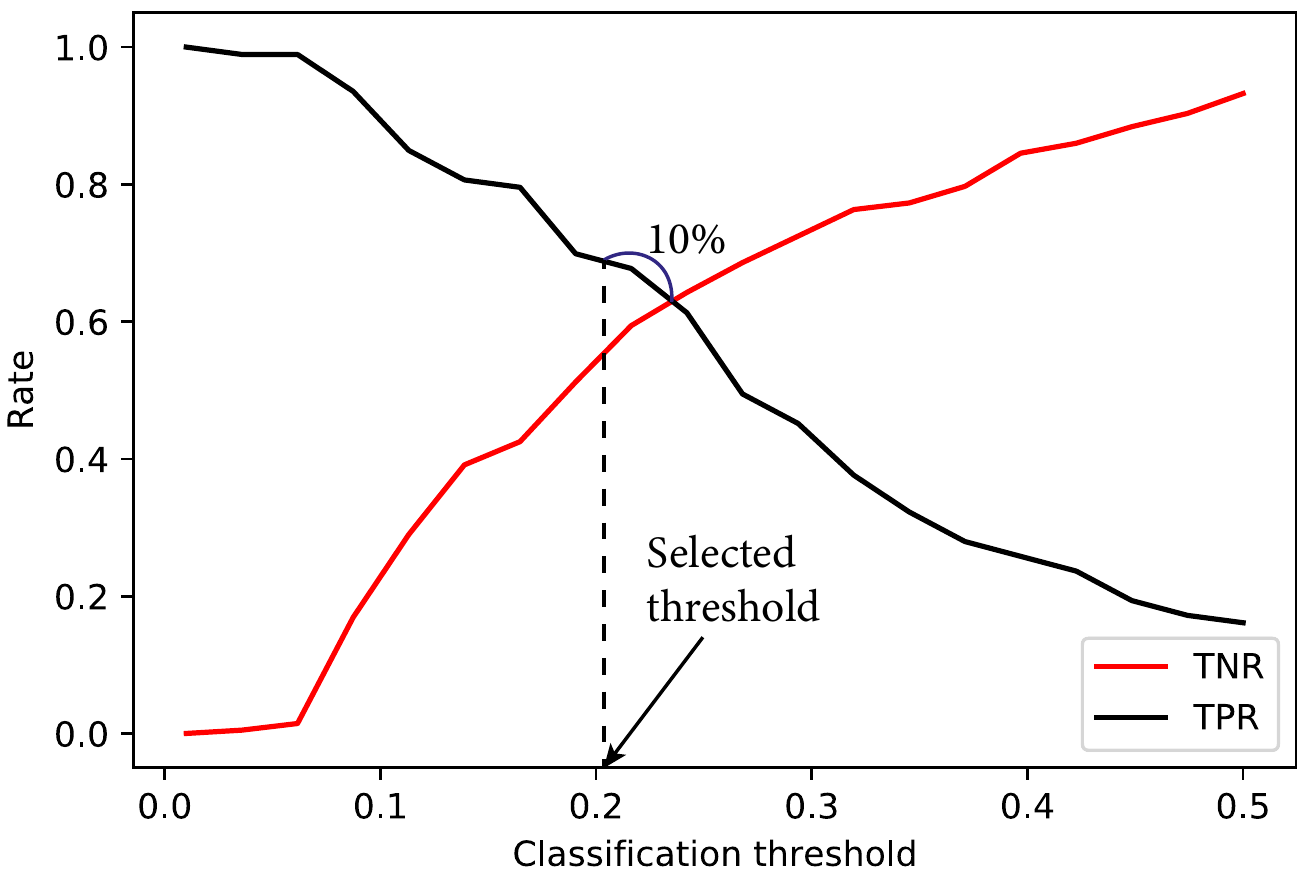} }}%
    \caption{Dependence of rates (TPR and TNR) from the classification threshold. The threshold is chosen as a 10\% offset from the point of intersection between TPR and TNR curves}%
    \label{rates}
\end{figure}
\setlength{\parindent}{0cm}In our case first-rate error is most critical because of this we pay special attention to TPR, but too high TPR will cause all individuals to be calculated according to the real model, thus TNR must also be taken into account. Therefore, we retreat 10\% to the left of the intersection point of the TPR/TNR curves.
\subsection{Comparison of optimization approaches}
\setlength{\parindent}{0.3cm}The main experiment is based on a comparison of four considered optimization approaches. Figure~\ref{HV} demonstrates convergence of the hypervolume for archive population during the iterative optimization.
\begin{figure}
    \centering
    {{\includegraphics[width=8.5cm]{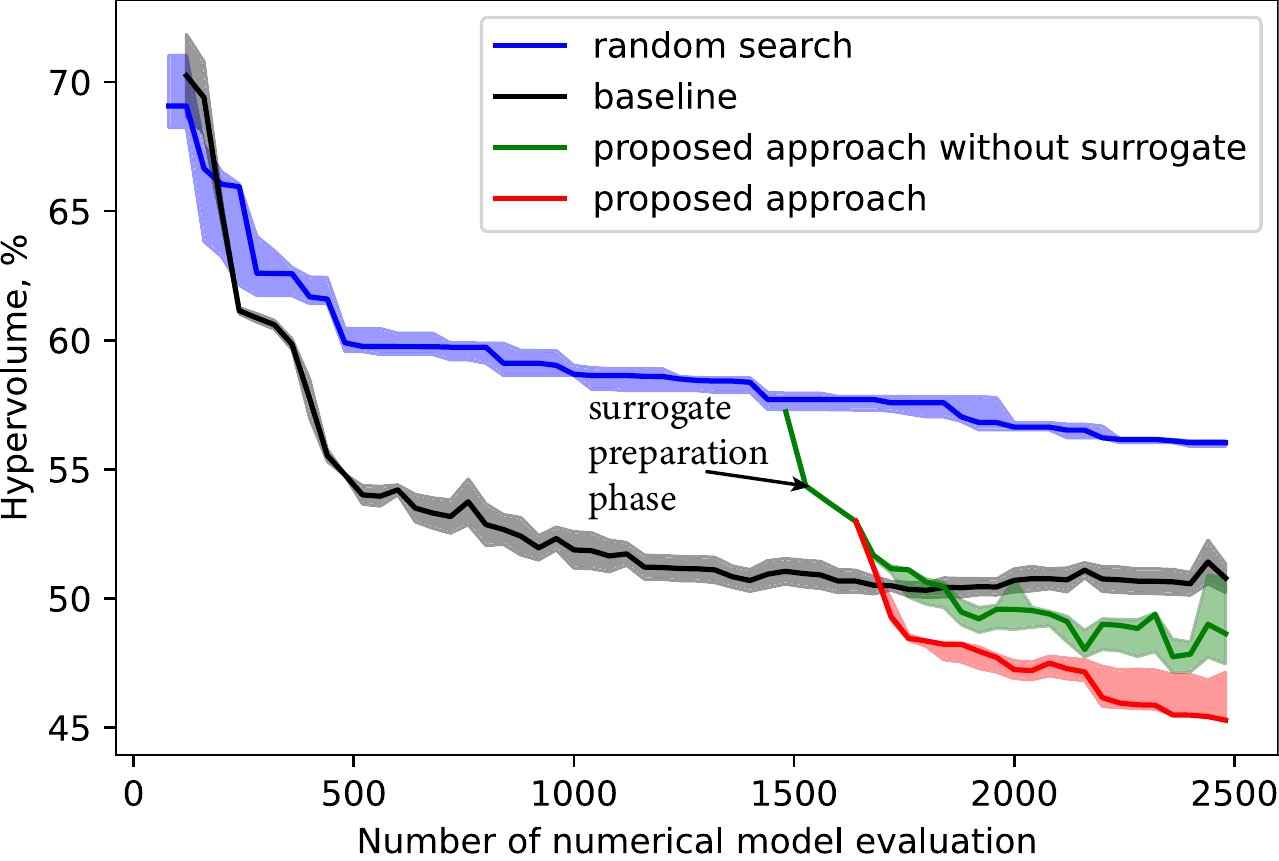} }}%
    \caption{Hypervolume convergence comparison of the four optimization approaches: random search, the baseline that is based on the SPEA2 evolutionary approach, and proposed approaches (with and without surrogate model).}%
    \label{HV}
\end{figure}
Each optimization process was run five times. For each iteration, 25th, 50th, 75th quantiles were calculated. As can be seen from Figure~\ref{HV}, proposed approaches start the evolutionary part at the point with 1500 real model evaluations, while the baseline and random search starts from 0. This is related to the proposed approaches using a large initialization stage with a population of size 1500, after which 40 best individuals are selected, and the evolutionary stage starts. At the beginning of the evolutionary stage, a real model is used since the computational effort for the preparation of the surrogate model (train surrogate and assistant, classification threshold selection) is roughly comparable to the time required to 160 evaluation of the numerical model. After the completion of the preparation phase real model is replaced by a surrogate model. As can be seen from the Figure~\ref{HV}, the proposed surrogate-assisted approach shows the best performance. The importance of the surrogate model and the large initialization stage is indicated by the higher hypervolume for the proposed approach without surrogate and baseline. In Figure~\ref{pf} the fittest individuals obtained during proposed optimization processes are shown. It can be seen that the surrogate algorithm gives better solutions than the non-surrogate approach. Individuals are closer to the lower-left corner and dominate individuals from a real model-based approach.
\begin{figure}
    \centering
    {{\includegraphics[width=9cm]{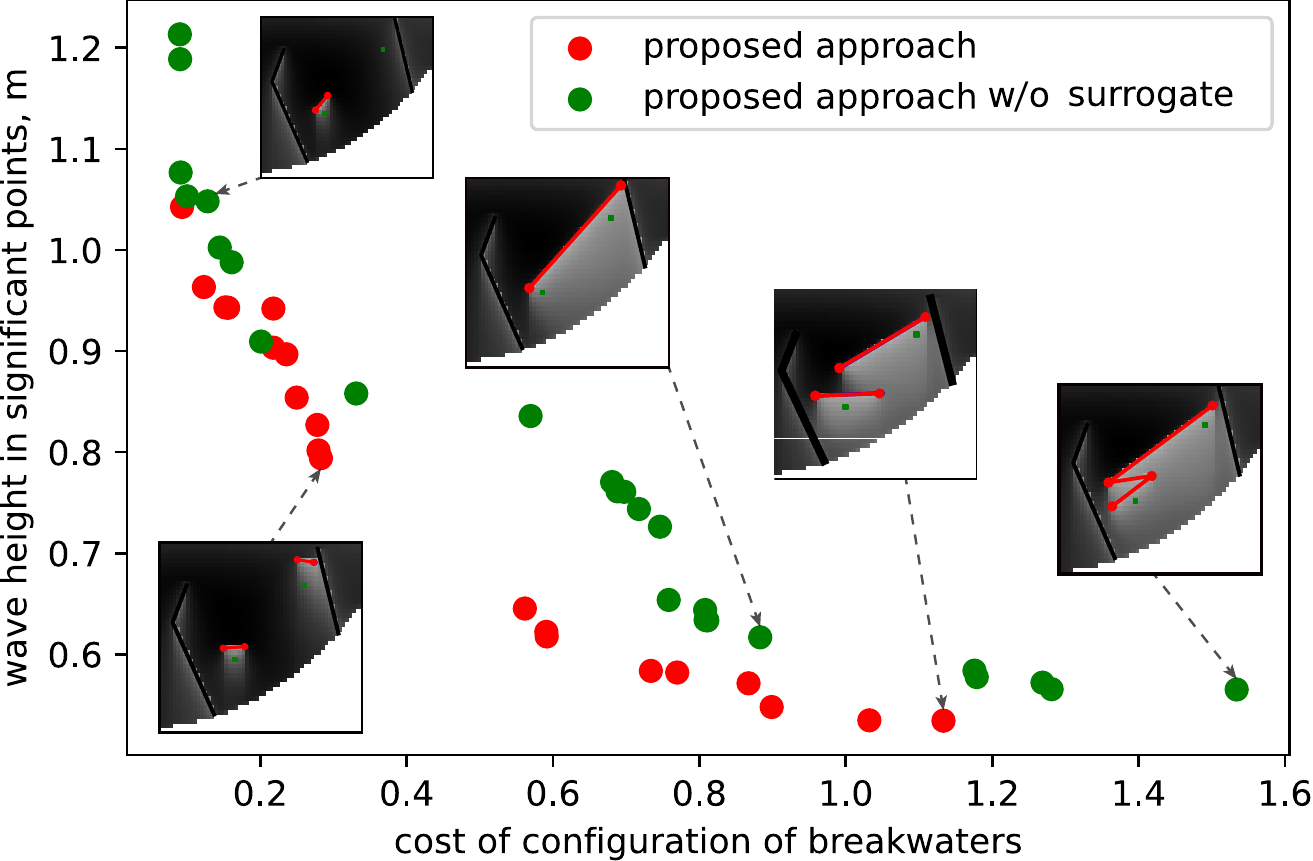} }}%
    \caption{Comparison between the best individuals (archive population) for proposed approaches with surrogate and real model.}%
    \label{pf}
\end{figure}
Table \ref{tab_dataset_features} presents the results of four approaches.
50th quantiles were taken as hypervolume values. The efficiency of cost and wave heights at important points was calculated from the baseline using the following formulas:
\begin{equation} \label{eq2}
\begin{gathered}
    cost \% = \frac{1}{S_{arch}}\sum_{wh}\left( \left( \frac{cost_{pred}}{cost_{baseline}} \right)_{wh}-1 \right)\cdot100
\end{gathered}
\end{equation}
\begin{equation} \label{eq3}
\begin{gathered}
    wh \% = \frac{1}{S_{arch}}\sum_{cost}\left( \left( \frac{wh_{pred}}{wh_{baseline}} \right)_{cost}-1 \right)\cdot100
\end{gathered}
\end{equation}
The proposed approach achieves improvements of 42\% in terms of cost and 17\%  in terms of wave defense effectiveness compared to the baseline, which empirically confirms \hyperref[hypmo]{hypothesis} for the considered coastal area.
\begin{table}[H]
\centering
\caption{Comparison between different optimization approaches. HV - hypervolume, WH - wave height.}
\label{tab_dataset_features}
\begin{tabular}{|c|c|c|c|} 
\hline
Method      & HV, \( \% \) & Cost, \( \% \) & WH, \( \% \) \\ 
\hline
random search        & 56          & +131             & +14  \\ 
\hline
baseline          & 51          & {-}             & {-} \\ 
\hline
prop. appr. w/o surrogate & $48$          & {$-21$}            & +0.002 \\ 
\hline
proposed approach      & $\boldsymbol{45}$           &$\boldsymbol{-42}$            & $\boldsymbol{-17}$    \\
\hline
\end{tabular}
\end{table}
\section{Conclusion}
\setlength{\parindent}{0.3cm}In this paper, we propose the surrogate-assisted multi-objective evolutionary approach for the generative design of breakwaters. The algorithm takes into account different numbers of breakwaters and their segments due to the unified representation in the form of three-dimensional parameterization. 

\setlength{\parindent}{0.3cm}Our approach is based on a surrogate model consisting of two neural networks - surrogate model and assistant. Both of them are convolutional neural networks. The assistant model was developed to estimate the confidence of predictions. To preserve promising solution, additional verification stage for archive individuals was added.

The results of the experiments confirm that the proposed approach 
provides better solutions for the same time than the non-surrogate approaches. The proposed approach achieves improvements of 42\% in terms of cost and 17\%  in terms of wave defense effectiveness compared to the evolutionary baseline. 

Future extensions of this research can be done by applying of generative models (generative adversarial networks or variational autoencoders) to produce the candidate solutions.

\section{Code and data availability}

The data and code for all experiments are available in the \url{https://github.com/quickjkee/generative-design-breakwaters} repository.

\section{Acknowledgements}

This work was supported by the Analytical Center for the Government of the Russian Federation (IGK 000000D730321P5Q0002), agreement No. 70-2021-00141.


\bibliographystyle{IEEEtran}
\bibliography{bibliography.bib}

\end{document}